\documentclass[11pt,a4paper]{article}
\usepackage{acl2017}
\usepackage{times}
\usepackage{multirow}
\usepackage{url}
\usepackage{latexsym}
\usepackage{graphicx}
\usepackage{color}
\usepackage{booktabs}

\aclfinalcopy % Uncomment this line for the final submission
\title{OpenNMT: Open-Source Toolkit for Neural Machine Translation}

\author{Guillaume Klein$^\dagger$, Yoon Kim$^*$, Yuntian Deng$^*$, Jean Senellart$^\dagger$, Alexander M. Rush$^*$ \\ Harvard University$^*$, SYSTRAN $^\dagger$}

\date{}

\begin{document}
\maketitle
\begin{abstract}

  We describe an open-source toolkit for neural machine translation
  (NMT).  The toolkit prioritizes efficiency, modularity, and
  extensibility with the goal of supporting NMT research into model
  architectures, feature representations, and source modalities, while
  maintaining competitive performance and reasonable training
  requirements. The toolkit consists of modeling and translation support,
  as well as detailed pedagogical documentation about the underlying
  techniques.

\end{abstract}

\section{Introduction}

Neural machine translation (NMT) is a new methodology for machine
translation that has led to remarkable improvements, particularly in
terms of human evaluation, compared to rule-based
and statistical machine translation (SMT) systems
\cite{wu2016google,systran}. Originally developed using pure
sequence-to-sequence models \cite{sutskever14sequence,Cho2014} and
improved upon using attention-based variants \cite{Bahdanau2015,Luong2015}, NMT has now become a widely-applied technique for machine
translation, as well as an effective approach for other related NLP
tasks such as dialogue, parsing, and summarization.

As NMT approaches are standardized, it becomes more important for the
machine translation and NLP community to develop open implementations
for researchers to benchmark against, learn from, and extend
upon. Just as the SMT community benefited greatly from toolkits like
Moses \cite{koehn2007moses} for phrase-based SMT and CDec
\cite{dyer2010cdec} for syntax-based SMT, NMT toolkits can provide a
foundation to build upon. A toolkit should aim to provide
a shared framework for developing and comparing open-source systems,
while at the same time being efficient and accurate enough to be used
in production contexts.

\begin{figure}
  \centering
  \includegraphics[width=\linewidth]{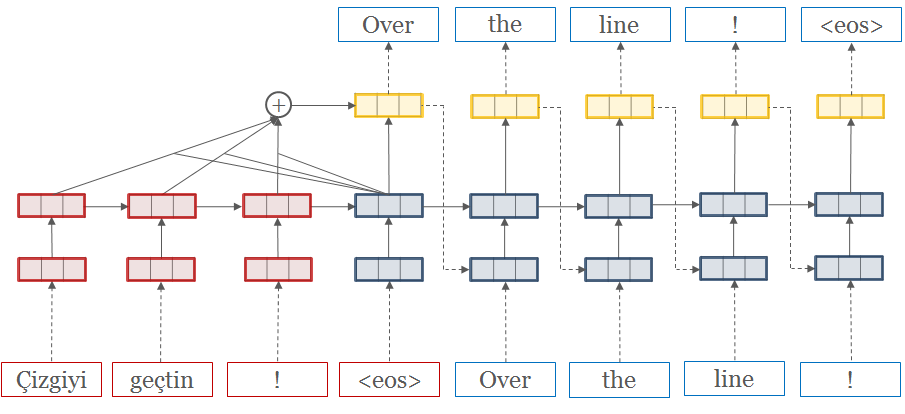}
  \label{fig:rnn}
  \caption{\small Schematic view of neural machine translation. The \textcolor{red}{red} source words are first mapped to word vectors and then fed into a recurrent neural network (RNN). Upon seeing the $\langle$eos$\rangle$ symbol, the final time step initializes a target \textcolor{blue}{blue} RNN. At each target time step, \textit{attention} is applied over the source RNN and combined with the current hidden state to produce a prediction $p(w_t| w_{1: t-1}, x)$ of the next word. This prediction is then fed back into the target RNN.}
\end{figure}

% In this work we describe a new open-source toolkit for developing
% neural machine translation systems, known as \textit{OpenNMT}. The
% system is motivated by frameworks, such as Moses and CDec developed
% for statistical machine translation (SMT). These toolkits aim to
% provide a shared frameworks for developing and comparing open-source
% SMT systems that are complete and flexible enough for research
% development, while at the same time being efficient and accurate
% enough to be used production contexts. 

Currently there are several existing NMT implementations. Many systems
such as those developed in industry by Google, Microsoft, and Baidu,
are closed source, and are unlikely to be released with unrestricted
licenses. Many other systems such as \textit{GroundHog},
\textit{Blocks}, \textit{tensorflow-seq2seq}, \textit{lamtram}, and
our own \textit{seq2seq-attn}, exist mostly as research code. These
libraries provide important functionality but minimal support to
production users. Perhaps most promising is the University of
Edinburgh's \textit{Nematus} system originally based on NYU's NMT
system. Nematus provides high-accuracy translation, many options,
clear documentation, and has been used in several successful research
projects. In the development of this project, we aimed to build upon
the strengths of this system, while providing additional documentation
and functionality to provide a useful open-source NMT framework for
the NLP community in academia and industry.

With these goals in mind, we introduce \textit{OpenNMT}
(\url{http://opennmt.net}), an open-source framework for neural
machine translation. OpenNMT is a complete NMT implementation. In
addition to providing code for the core translation tasks, OpenNMT was
designed with three aims: (a) prioritize first training and test
efficiency, (b) maintain model modularity and readability, (c) support
significant research extensibility.

This engineering report describes how the system targets these
criteria. We begin by briefly surveying the background for NMT,
describing the high-level implementation details, and then describing
specific case studies for the three criteria.  We end by showing
benchmarks of the system in terms of accuracy, speed, and memory usage
for several translation and translation-like tasks.

% the main aspects of system development, and the preliminary results from using the system in practice.  

% (a) Our top priority was training and decoding speed. NMT systems are
% notoriously slow to train, often requiring weeks of training time on
% the latest GPU hardware and significant test resources. We targeted
% this issue by implementing multi-GPU training, using aggressive memory
% sharing, and developing a specialized CPU decoder. 

% (b) Our second
% priority was system modularity and teachibility. We intend OpenNMT as
% a living research system, and so the codebase was developed to provide
% a self-documenting overview of NMT. We discuss how this approach
% allowed us to add factored models \cite{} to the system. 

% (c) Finally
% NMT is a very quick moving research area, and we would like the system
% to support new research areas as they develop. To demonstrate this
% approach we abstracted out the core of OpenNMT as a library, and
% describe the case study of using OpenNMT for image-to-text
% translation.

\section{Background}

NMT has now been extensively described in many
excellent tutorials (see for instance
\url{https://sites.google.com/site/acl16nmt/home}). We give only
a condensed overview. 

NMT takes a conditional language modeling view of translation by modeling the
probability of a target sentence $w_{1:T}$ given a source sentence
$x_{1:S}$ as
$p(w_{1:T}| x) = \prod_{1}^T p(w_t| w_{1:t-1}, x; \theta)$. This
distribution is estimated using an attention-based encoder-decoder
architecture \cite{Bahdanau2015}. A source encoder recurrent neural
network (RNN) maps each source word to a word vector, and processes
these to a sequence of hidden vectors
$\mathbf{h}_1, \ldots, \mathbf{h}_S$.  The target decoder combines an
RNN hidden representation of previously generated words
($w_1, ... w_{t-1}$) with source hidden vectors to predict scores for
each possible next word. A softmax layer is then used to produce a
next-word distribution $ p(w_t| w_{1:t-1}, x; \theta)$. The source
hidden vectors influence the distribution through an attention pooling
layer that weights each source word relative to its expected
contribution to the target prediction. The complete model is trained
end-to-end to minimize the negative log-likelihood of the training
corpus. An unfolded network diagram is shown in Figure~\ref{fig:rnn}.

In practice, there are also many other important aspects that improve
the effectiveness of the base model. Here we briefly mention four
areas: (a) It is important to use a gated RNN such as an LSTM
\cite{hochreiter1997long} or GRU \cite{chung2014empirical} which help
the model learn long-term features. (b) Translation requires
relatively large, stacked RNNs, which consist of several vertical
layers (2-16) of RNNs at each time step \cite{sutskever14sequence}. (c)
Input feeding, where the previous attention vector is fed back into
the input as well as the predicted word, has been shown to be quite
helpful for machine translation \cite{Luong2015}.  (d) Test-time
decoding is done through \textit{beam search} where multiple
hypothesis target predictions are considered at each time
step. Implementing these correctly can be difficult, which motivates
their inclusion in an NMT framework.

% \begin{itemize}
% \item One column describing the technical details
% \end{itemize}

\section{Implementation}

OpenNMT is a complete library for training and deploying neural
machine translation models. The system is successor to
\textit{seq2seq-attn} developed at Harvard, and has been completely
rewritten for ease of efficiency, readability, and
generalizability. It includes vanilla NMT models along with support
for attention, gating, stacking, input feeding, regularization, beam
search and all other options necessary for state-of-the-art
performance.  

The main system is implemented in the Lua/Torch mathematical
framework, and can be easily be extended using Torch's internal
standard neural network components. It has also been extended by Adam
Lerer of Facebook Research to support Python/PyTorch framework, 
with the same API.

\begin{figure}
  \centering
  \includegraphics[width=\linewidth]{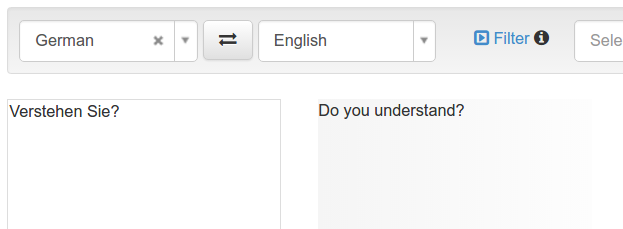}
  \label{fig:live}
  \caption{\small Live demo of the OpenNMT system across dozens of language pairs.}
\end{figure}

The system has been developed completely in the open on GitHub at
(\url{http://github.com/opennmt/opennmt}) and is MIT licensed.  The
first version has primarily (intercontinental) contributions from
SYSTRAN Paris and the Harvard NLP group. Since official beta release,
the project has been starred by over 1000 users, and there have been
active development by those outside of these two organizations. The
project has an active forum for community feedback with over five
hundred posts in the last two months. There is also a live
demonstration available of the system in use (Figure~\ref{fig:live}).

One nice aspect of NMT as a model is its relative compactness. The Lua
OpenNMT system including preprocessing is roughly 4K lines of code,
and the Python version is less than 1K lines (although slightly
less-feature complete). For comparison the Moses SMT framework
including language modeling is over 100K lines. This makes the system
easy to completely understand for newcomers. The project is fully
self-contained depending on minimal number of external Lua libraries
and including also a simple language independent reversible
tokenization and detokenization tools.

\section{Design Goals}

As the low-level details of NMT have been covered previously, we focus
this report on the design goals of OpenNMT: system efficiency, code modularity, and model
extensibility.

\subsection{System Efficiency}

As NMT systems can take from days to weeks to train, training
efficiency is a paramount concern. Slightly faster training can make be the difference between
plausible and impossible experiments.

\paragraph{Memory Sharing}

When training GPU-based NMT models, memory size restrictions are the
most common limiter of batch size, and thus directly impact training
time. Neural network toolkits, such as Torch, are often designed to
trade-off extra memory allocations for speed and declarative
simplicity. For OpenNMT, we wanted to have it both ways, and so we
implemented an external memory sharing system that exploits the known
time-series control flow of NMT systems and aggressively shares the
internal buffers between clones. The potential shared buffers are
dynamically calculated by exploration of the network graph before
starting training. In practical use, aggressive memory reuse provides
a saving of 70\% of GPU memory with the default model size.

\paragraph{Multi-GPU} OpenNMT additionally supports multi-GPU training
using data parallelism. Each GPU has a replica of the master
parameters and process independent batches during training phase.  Two
modes are available: synchronous and asynchronous training.  In
synchronous training, batches on parallel GPU are run simultaneously
and gradients aggregated to update master parameters before
resynchronization on each GPU for the following batch.  In
asynchronous training, batches are run independent on each GPU, and
independent gradients accumulated to the master copy of the
parameters. Asynchronous SGD is known to provide faster convergence
\cite{dean2012large}. Experiments with 8 GPUs show a 6$\times$ speed up in 
per epoch, but a slight loss in training efficiency. When training to similar
loss, it gives a 3.5$\times$ total speed-up to training.

\paragraph{C/Mobile/GPU Translation} Training NMT systems requires
significant code complexity to facilitate fast
back-propagation-through-time. At deployment, the system is much less
complex, and only requires (i) forwarding values through the network
and (ii) running a beam search that is much simplified compared to
SMT. OpenNMT includes several different translation deployments
specialized for different run-time environments: a batched CPU/GPU
implementation for very quickly translating a large set of sentences,
a simple single-instance implementation for use on mobile devices, and
a specialized C implementation. The first implementation is suited for
research use, for instance allowing the user to easily include
constraints on the feasible set of sentences and ideas such as pointer
networks and copy mechanisms. The last implementation is particularly
suited for industrial use as it can run on CPU in standard production
environments; it reads the structure of the network and then uses the
\textit{Eigen} package to implement the basic linear algebra necessary
for decoding. Table~\ref{tab:cpu} compares the performance of the
different implementations based on batch size, beam size.

\begin{table}
  \centering
  \begin{tabular}{ccccc}
    \toprule
    Batch & Beam & GPU & CPU & CPU/C \\ 
    \midrule
    1  & 5 & 209.0 & 24.1 & 62.2\\
    1  & 1 & 166.9 & 23.3 & 84.9\\
    30 & 5 & 646.8 & 104.0 & 116.2\\
    30 & 1 & 535.1 & 128.5  & 392.7\\

    \bottomrule
  \end{tabular}

  \label{tab:cpu}
  \caption{\small Performance numbers in source tokens per second for the Torch CPU/GPU implementations and for 
  the  multi-threaded CPU C implementation. (Run with Intel i7/GTX 1080)}
\end{table}

\subsection{Modularity for Research}

A secondary goal was a desire for code readability for non-experts.
We targeted this goal by explicitly separating out many optimizations
from the core model, and by including tutorial documentation within
the code. To test whether this approach would allow novel feature
development we experimented with two case studies.

\paragraph{Case Study: Factored Neural Translation}

In feature-based factored neural translation
\cite{sennrich2016linguistic}, instead of generating a word at each
time step, the model generates both word and associated features. For
instance, the system might include words and separate case features. This extension
requires modifying both the inputs and the output of the decoder to
generate multiple symbols. In OpenNMT both of these aspects are
abstracted from the core translation code, and therefore factored
translation simply modifies the input network to instead process the
feature-based representation, and the output generator network to
instead produce multiple conditionally independent predictions.

\paragraph{Case Study: Attention Networks}

The use of attention over the encoder at each step of translation is
crucial for the model to perform well. The default method is to
utilize the global attention mechanism. However
there are many other types of attention that have recently proposed
including local attention \cite{Luong2015}, sparse-max attention
\cite{martins2016softmax}, hierarchical attention
\cite{yang2016hierarchical} among others. As this is simply a module
in OpenNMT it can easily be substituted. Recently the Harvard
group developed a \textit{structured} attention approach,
that utilizes graphical model inference to compute this attention. The
method is quite computationally complex; however as it is modularized by the Torch
interface, it can be used in OpenNMT to substitute
for standard attention.

\subsection{Extensibility}

Deep learning is a quickly evolving field. Recently work such as
variational seq2seq auto-encoders
\cite{DBLP:conf/conll/BowmanVVDJB16} or memory networks
\cite{DBLP:journals/corr/WestonCB14}, propose interesting extensions
to basic seq2seq models. We next discuss a case study to demonstrate that OpenNMT
is extensible to future variants.

% A final goal of OpenNMT is  the deep learning is a very
% quick moving area and that likely within the next couple years there
% will be many unexpected applications of these style of
% methods. 

\paragraph{Multiple Modalities}

Recent work has shown that NMT-like systems are effective for
image-to-text generation tasks
\cite{DBLP:journals/corr/XuBKCCSZB15}. This task is quite different
from standard machine translation as the source sentence is now an
image.  However, the future of translation may require this style of
(multi-)modal inputs
(e.g. \url{http://www.statmt.org/wmt16/multimodal-task.html}).  

As a case study, we adapted two systems with non-textual inputs to run
in OpenNMT. The first is an image-to-text system developed for
mathematical OCR \cite{DBLP:journals/corr/DengKR16}.  This model
replaces the source RNN with a deep convolution over the source
input. Excepting preprocessing, the entire adaptation requires less
than 500 lines of additional code and is also open-sourced as
\url{github.com/opennmt/im2text}.  The second is a speech-to-text recognition
system based on the work of \citet{DBLP:journals/corr/ChanJLV15}.
This system has been implemented directly in OpenNMT by replacing the
source encoder with a Pyrimidal source model.

\subsection{Additional Tools}

\begin{figure}
  \centering
  \includegraphics[width=\linewidth]{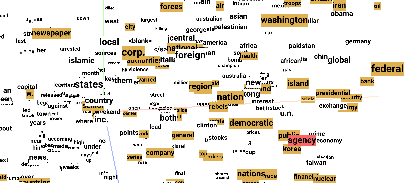}
  \label{fig:tensor}
  \caption{\small 3D Visualization of OpenNMT source embedding from the TensorBoard visualization system.}
\end{figure}

Finally we briefly summarize some of the additional tools that extend
OpenNMT to make it more beneficial to the research community.

\begin{table*}
   \centering
  \begin{tabular}{lccccc}
    \toprule
          & ES & FR & IT & PT & RO \\
    \midrule
ES	& - &  32.7 (+5.4)	 &  28.0 (+4.6) &  34.4 (+6.1) &  28.7 (+6.4) \\ 
FR	&  32.9 (+3.3)	& -  &  26.3 (+4.3)	 &  30.9 (+5.2) &  26.0 (+6.6) \\
IT     &  31.6 (+5.3)	 &  31.0 (+5.8) & - &  28.0 (+5.0) &  24.3 (+5.9) \\
PT	&  35.3 (+10.4) &  34.1 (+4.7) &  28.1 (+5.6) & - &  28.7 (+5.0) \\
RO	&  35.0 (+5.4) &  31.9 (+9.0) &  26.4 (+6.3) &  31.6 (+7.3) & -\\
    \bottomrule
  \end{tabular}

  \caption{  \label{tab:esfritptro}   20 language pair single translation model. Table shows BLEU($\Delta$) where $\Delta$ compares to only using the pair for training. }
\end{table*}

\begin{table}
  \centering
  \begin{tabular}{ccccc}
    \toprule
    Vocab & System & \multicolumn{2}{c}{Speed tok/sec}  & BLEU\\
     &  & Train  & Trans  &  \\
    \midrule
 V=50k & Nematus  & 3393 & 284 & 17.28 \\
     & ONMT  &4185 & 380 & 17.60 \\ 
    \midrule 
  V=32k & Nematus & 3221& 252 & 18.25 \\
    & ONMT &5254 & 457 & 19.34\\ 
    \bottomrule
  \end{tabular}

  \caption{ \small \label{tab:res} Performance Results for EN$\rightarrow$DE on WMT15 tested on \textit{newstest2014}. Both system 2x500 RNN, embedding size 300, 13 epochs, batch size 64, beam size 5. We compare on a 50k vocabulary and a 32k BPE setting.}
\end{table}

\paragraph{Tokenization} We aimed for OpenNMT to be a standalone project
and not depend on commonly used tools.  For instance the Moses
tokenizer has language specific heuristics not necessary in NMT. We
therefore include a simple reversible tokenizer that (a) includes
markers seen by the model that allow simple deterministic
detokenization, (b) has extremely simple, language-independent
tokenization rules. The tokenizer can also perform Byte Pair Encoding
(BPE) which has become a popular method for sub-word tokenization
 in NMT systems \cite{DBLP:journals/corr/SennrichHB15}.

 \paragraph{Word Embeddings} OpenNMT includes tools for simplifying
 the process of using pretrained word embeddings, even allowing
 automatic download of embeddings for many languages. This allows
training in languages or domain with relatively little aligned data.
Additionally OpenNMT can export the word embeddings from trained 
models to standard formats. This allows analysis is external tools 
such as TensorBoard, shown in Figure~\ref{fig:tensor}.

\section{Benchmarks}

We now document some runs of the model. We expect performance and
memory usage to improve with further development.  Public benchmarks
are available at \url{http://opennmt.net/Models/}, which also includes
publicly available pre-trained models for all of these tasks and
tutorial instructions for all of these tasks. The benchmarks are
run on a Intel(R) Core(TM) i7-5930K CPU @ 3.50GHz, 256GB Mem,
trained on 1 GPU GeForce GTX 1080 (Pascal) with CUDA v. 8.0 (driver
375.20) and cuDNN (v. 5005).

The comparison, shown in Table~\ref{tab:res}, is on English-to-German
(EN$\rightarrow$DE) using the WMT
2015\footnote{\url{http://statmt.org/wmt15}} dataset. Here we compare,
BLEU score, as well as training and test speed to the publicly
available \textit{Nematus} system. 
\footnote{\url{https://github.com/rsennrich/nematus}. Comparison with
  OpenNMT/Nematus github revisions {\tt 907824}/{\tt 75c6ab1}.}

We additionally trained a multilingual translation model following
\newcite{johnson2016google}. The model translates from and to French,
Spanish, Portuguese, Italian, and Romanian. Training data is 4M
sentences and was selected from the open parallel
corpus\footnote{\url{http://opus.lingfil.uu.se}}, specifically from
Europarl, GlobalVoices and Ted. Corpus was selected to be
multi-source, multi-target: each sentence has its translation in the 4
other languages. Corpus was tokenized using shared Byte
Pair Encoding of 32k.  Comparative results between multi-way
translation and each of the 20 independent training are presented in
Table~\ref{tab:esfritptro}. The systematically large improvement shows
that language pair benefits from training jointly with the other language pairs.

Additionally we have found interest from the community in using
OpenNMT for non-standard MT tasks like sentence document summarization
dialogue response generation (chatbots), among others.  Using
OpenNMT, we were able to replicate the sentence summarization results
of \citet{chopra2016abstractive}, reaching a ROUGE-1 score of 33.13 on
the Gigaword data. We have also trained a model on 14 million
sentences of the OpenSubtitles data set based on the work
\citet{vinyals2015neural}, achieving comparable perplexity.

\section{Conclusion}

We introduce \textit{OpenNMT}, a research toolkit for NMT that
prioritizes efficiency and modularity. We hope to further develop
OpenNMT to maintain strong MT results at the research frontier, 
providing a stable and framework for production use.

\newpage

\bibliography{writeup}
\bibliographystyle{acl_natbib}

\end{document}